\documentclass[conference]{IEEEtran}
\IEEEoverridecommandlockouts
\usepackage{cite}
\usepackage{amsmath,amssymb,amsfonts}
\usepackage{algorithmic}
\usepackage{graphicx}
\usepackage{subcaption} 
\usepackage{textcomp}
\usepackage{xcolor}
\usepackage{booktabs}
\usepackage{algorithm}
\usepackage{algorithmic}
\usepackage{amsmath}
\usepackage{multirow}
\def\BibTeX{{\rm B\kern-.05em{\sc i\kern-.025em b}\kern-.08em
    T\kern-.1667em\lower.7ex\hbox{E}\kern-.125emX}}
\begin{document}

\title{Effective Unbounded Drift Control in Heterogeneous Federated Learning: Bayesian Fine-Tuning Using a Reference Model\\
}

\author{
	\IEEEauthorblockN{Taehwan Yoon\IEEEauthorrefmark{1}, Wesley De Neve\IEEEauthorrefmark{2}, Bongjun Choi\IEEEauthorrefmark{3}}
\IEEEauthorblockA{\IEEEauthorrefmark{1}\IEEEauthorrefmark{3}\textit{School of Computer Science and Engineering}, 
\textit{Soongsil University}, 
Seoul, Republic of Korea \\
Email: \{dbs1045, davidchoi\}@soongsil.ac.kr}
\IEEEauthorblockA{\IEEEauthorrefmark{2}\textit{Center for Biosystems and Biotech Data Science,} 
	\textit{Ghent University}, 
	Global Campus, Incheon, Republic of Korea \\
	Email: wesley.deneve@ghent.ac.kr}
}

\maketitle

\begin{abstract}
Federated learning (FL) enables collaborative model training across distributed clients while preserving data privacy. However, data and system heterogeneity often induce unbounded drift in model updates due to each clients' different multiple local steps, partial participation, and unbounded dynamics in model updates. These challenges severely degrade predictive performance, exacerbate catastrophic forgetting, and hinder global optimization forward to optimal updates. Related works (AdaBest, FedEve, A-FedPD) defined dual drift problems and proposed each regularization term to client optimization process. However, in aspect of temporal unbounded drift, they generally reflect last round temporal drift to client optimization while forgets prior temporal unbounded drift. To address these limitations, we propose FedRef, a robust optimization framework that leverages a reference model generalized from prior global models via Bayesian fine-tuning. FedRef integrates a Maximum A Posteriori (MAP)-based regularization mechanism that calibrates global updates toward a temporally aggregated reference model, thereby effectively controlling unbounded drift and maximizing empirical stability.
Moreover, unlike prior approaches that rely on complex client-side optimization heuristics, FedRef achieves superior generalization to reflect recency weighting by seamlessly blending a robust probabilistic prior with global likelihood optimization. By maintaining a temporal moving average centered on previous global trajectories, the proposed framework provides a highly reliable optimization anchor that effectively counteracts misleading updates under extreme unbounded drift settings. Extensive experiments on image classification (FEMNIST, CINIC-10) and medical image segmentation (FeTS2022) demonstrate that FedRef yields superior predictive performance, and higher F1-scores, even under severe non-IID conditions. These results highlight FedRef as an exceptionally efficient and stable framework to address unbounded drift in model updates and achieve high-fidelity generalization including prior round global models' available information in heterogeneous real-world FL scenarios.
\end{abstract}


\section{Introduction}

Federated learning (FL) has recently been proposed as a promising solution to protect user data privacy while allowing collaborative model training between independent clients (devices or institutions)~\cite{yang2019federated,lim2020federated, fedavg}. User data is fundamentally protected because clients share only updated local model parameters (resulting from local training using local user data) and other non-private values with the server or other clients. The server then computes the global model by aggregating the local model updates received from the clients. However, FL approaches face numerous challenges, as follows:

\begin{itemize}
\item Statistical heterogeneity: this problem includes cases of skew distributions, data imbalance, model divergence, misleading or malicious clients behaviors (data poisoning).
\item Communication overhead: the cases of massive models (large models), bandwidth limitations, synchronous model updates can lead to communication overhead problems.
\item Clients' computational burden: computations of model optimization, roles of specific actions or connections, device constraints can increase clients' computational burden.
\item Security \& privacy risks: we need defenses of reconstruction attacks, model poisoning, model spoofing, and diverse attack scenarios.
\end{itemize}

Existing studies have introduced regularization and fine-tuning approaches to mitigate unbounded drift caused by malicious or misleading client behavior~\cite{babendererde2025jointly, dupuy2023quantifying}. 

Unbounded drift in federated learning arises mainly from the following three factors:
\begin{enumerate}
    \item Multiple local steps: performing multiple local optimization steps can cause model overfitting and drift away from the global model~\cite{varno2022adabest}.
    \item Partial participation: if the participating clients have data distributions that differ substantially from those of the overall client population, the global objective may shift from round to round, causing instability \cite{shen2025fedeve}.
    \item Unbounded dynamics: under relaxed smoothness and extreme heterogeneity, the distance between consecutive global models cannot be trivially bounded, leading to unaligned dual-drift and theoretical divergence \cite{sun2024fedpd}.
\end{enumerate}

As a result, unbounded drift in federated learning can lead to severe performance degradation, catastrophic forgetting, and inefficient client-side computation.
Our main objective is to mitigate unbounded drift in model updates without increasing client-side computational overhead.

Representative approaches for handling heterogeneity in FL, such as FedProx~\cite{fedprox} and various adaptive variants~\cite{reddi2020adaptive,varno2022adabest,shen2025fedeve}, primarily rely on enforcing explicit regularization terms or bias estimation updates on the client side. While these strategies attempt to constrain model divergence, they often introduce rigid optimization heuristics that can confuse participating clients and lead clients to low-performing participants. Under extreme statistical heterogeneity and unbounded dynamics, such restrictive local objectives inevitably hinder the global model from capturing the true underlying data distribution, thereby leading to severe generalization barriers, suboptimal predictive performance, and degraded convergence stability. 

To overcome these limitations and achieve robust global optimization, we introduce FedRef, a novel framework designed to enhance generalization and effective unbounded drift control via reference-model-based Bayesian fine-tuning. Moving away from rigid, ad-hoc constraints on client updates, FedRef formalizes a rigorous Maximum A Posteriori (MAP) estimation framework that organically blends the collective global likelihood with a stable probabilistic prior. Specifically, by synthesizing a temporally aggregated reference model that extracts comprehensive knowledge from past global trajectories, our approach provides a resilient optimization anchor. This temporal calibration mechanism systematically penalizes erratic updates driven by extreme non-IID data or adversarial dynamics, guiding the global model toward a temporally stable and highly generic parameter space without empirical performance degradation.

In summary, our main contributions are as follows:
First, we formalize a robust Bayesian optimization framework, FedRef, for heterogeneous federated learning. We derive a rigorous Maximum A Posteriori (MAP) objective that calibrates global updates against a temporally aggregated reference model, effectively suppressing unbounded drift without introducing distortive client-side heuristics.
Second, we establish an advanced stabilization mechanism driven by temporal history. By introducing a temporal proximity weighting scheme over past global trajectories, our framework provides a highly reliable probabilistic prior that accelerates global consensus and achieves superior generalization capabilities under extreme non-IID and data poisoning environments.
Third, we empirically validate FedRef across diverse tasks, including computer vision (FEMNIST and CINIC-10) and medical image segmentation (FeTS2022) benchmarks. These evaluations show that FedRef consistently outperforms state-of-the-art baselines in terms of predictive accuracy, F1-scores, and convergence velocity.

\begin{table}[!hbt]
\caption{List of Symbols and Abbreviations}
\centering
\begin{tabular}{|c|l|}
\hline
\textbf{Symbol} &  \textbf{Description}\\
\hline
    $A$ & Aggregation function\\
    $s$ & Local training step \\
    $k$ & Client index, $k = \{1,2, \ldots,K\}$\\
    $\mathcal{B}^{s}_k$ &  Mini-batch sampled from local dataset of client $k$ \\
    $D_k$ & Dataset of client $k$\\
    $D_{ref}$ & Synthetic data of reference model\\
    $\delta$ & Unbounded drift penalty threshold\\
    $\eta$ & Learning rate on local training\\
    $F$  & Cost function \\
    FIM & Fisher Information Matrix\\
    $\theta^*$ & Theoretical optimal model parameter\\
    $\hat\theta$ & Model parameter estimated on Bayesian fine-tuning\\
    $\theta^s$ & Model parameter of step $s$\\
    $\theta_{k}$ & Model parameter of client $k$\\
    $\theta^g$ & Global model parameter\\
    $\theta^{ref}$ & Reference model parameter\\
    $\log p$ & Log-posteriori function\\
    $\lambda^g$ &  Regularization of a proposed term strength value \\
    $r$ & Communication round, $k=\{1, 2, \ldots, R\}$\\
    $\psi$ & Catastrophic forgetting measure\\
    $\phi$ & Sum of $\rho$ range $\sum_{i=1}^\rho i$\\
    UDP & Unbounded drift penalty\\
    $\rho$ & Number of global models from previous rounds\\
    $W_k$ & Aggregation weights of client $k$\\
    $H$ & Hessian matrix\\
    \hline
\end{tabular}\label{tab:symbols}
\end{table}

\begin{figure*}[!t]
\centering
\includegraphics[width=0.9\textwidth]{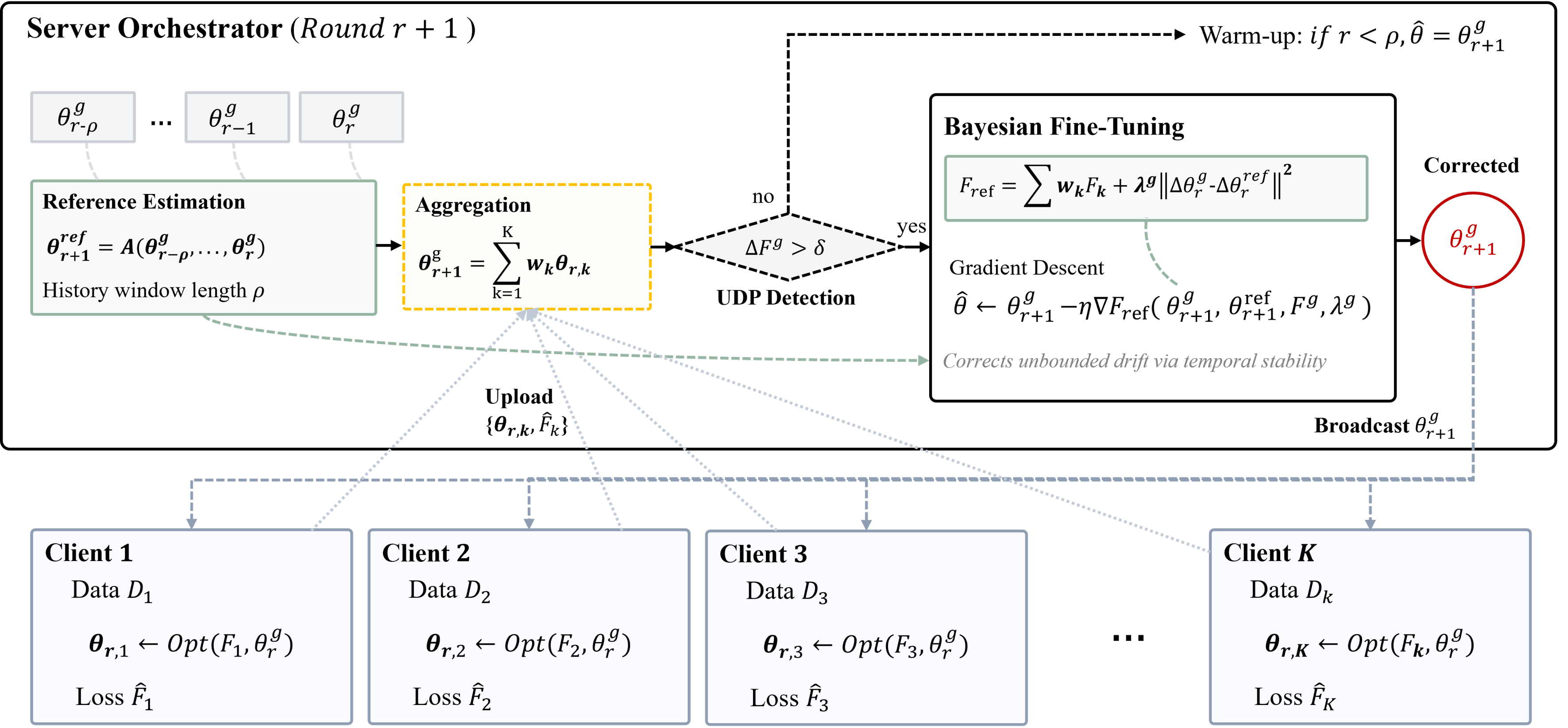}
\caption{Overview of our FedRef system.}
\label{fig:federatedlearningsystem}
\end{figure*}

\section{Related Work}

In this section, we present representative prior works on federated learning focused on model optimization and fine‑tuning for alleviating unbounded drift in model updates.


\subsection{FedAvg~\cite{fedavg}} is a baseline for federated learning. It based on stochastic gradient descent (SGD) \cite{chen2016revisiting} and aggregates local updates from clients to generate a global model $\theta^g$ by weighting them according to the number of samples per client, computed as follows:
\begin{equation}
\theta_{r+1}^g = \sum_{k=1}^K \frac{n_k}{n} \theta_{r,k},\label{averaging}
\end{equation}
where $r=\{1, 2, \ldots, R\}$ represents the communication round, $k=\{1, 2, \ldots, K\}$ represents the client index, $n_k$ represents the number of samples of client $k$, $n$ represents the total number of client samples, and $\theta_{r,k}$ represents the model parameter of client $k$ in round $r$. Compared with simple FedSGD aggregation, FedAvg generally provides more stable convergence and better empirical performance in federated settings.

On the client side, the local training process is updated in each training round using the loss function evaluated on the local test samples  for that round, as follows:
\begin{equation}
\theta_{r,k}^{s+1} = \theta_{r,k}^{s} - \eta \nabla F_k(\theta_{r,k}^{s}; \mathcal{B}_k^{s}),\label{localtraining}
\end{equation}
where $s$ represents the local training step, $\eta$ represents the learning rate, $\mathcal{B}_k^{s}$ represents the mini-batch sampled from the local data samples of client $k$ in step $s$, and $F_k(\theta_{k}^{s}; \mathcal{B}_{k}^{s})$ represents the local loss function evaluated on a mini-batch.


\subsection{FedProx~\cite{fedprox}} aims to optimize the predictive performance of federated learning under heterogeneous client conditions. A proximal term is introduced in the objective function of each client to ensure that local models do not deviate substantially from the global model, computed as follows:
\begin{equation}
\min_{\theta_k} \left(\mathbb{E}_{\mathcal{D}_{k}} \left[ \mathcal{L}(\theta_{r,k}, \mathcal{D}_k) \right] + \frac{\mu}{2} \| \theta_{r,k}^s - \theta^{s-1}_r \|^2 \right),\label{eq:fedprox}
\end{equation}
where $\mathcal{L}(\theta_k, \mathcal{D}_k)$ represents the cost function evaluated on the local data $\mathcal{D}_k$ of client $k$, and $\mu$ represents the hyperparameter that controls the proximal term strength. Although this approach improves robustness to data heterogeneity, it is hard to address unbounded drift issues. The server then simply aggregates the client models and sends the updated global model back to the clients.

\subsection{FedOpt~\cite{reddi2020adaptive}} incorporates adaptive optimization algorithms—such as Adam, Yogi, and Adagrad—on the server side, resulting in server optimized variants of FedAvg, namely FedAdam, FedYogi, and FedAdagrad. These methods, however, still require clients to compute additional regularization terms during local optimization, thereby increasing client‑side computational cost.
\begin{equation}
    \min_{x \in D_k} f(x) := \sum_{k=1}^{K} p_k F_k(x).\label{eq:fedopt}
\end{equation}

\subsection{AdaBest~\cite{varno2022adabest}} adaptively estimates client drift across heterogeneous clients to improve stability by constraining the norm of drift-related updates while reducing storage, communication, and computational overhead. They apply fine-tuning and optimization on both the client and server sides. They demonstrated better predictive performance on multi-label vision tasks than FedDyn~\cite{acar2021federateddyn} and SCAFFOLD~\cite{karimireddy2020scaffold}.
\begin{equation}
    \nabla F_{adabest} = \nabla F_{r,k} - h_{r,k},
    \label{eq:adabest}
\end{equation}
where $h_{r,k}$ represents the adaptive bias estimate utilized to mitigate client drift caused by data heterogeneity across clients.

\subsection{FedEve~\cite{shen2025fedeve}} addresses both period drift and client drift through a predict-observe framework that compensates for these two sources of drift and reduces their overall impact.
Their framework includes a Bayesian filtering approach that reduces variance relative to the individual variances associated with period drift and client drift.
\begin{equation}
    \nabla F_{eve} = \nabla F_{r,k} - \eta^g M_{r,k}+Q_{r,k} -\eta^g\Delta\theta_{r-1}+R_{r,k}, \label{eq:fedeve}
\end{equation}
where, $M_{r,k}$, $Q_{r,k}$, and $R_{r,k}$ denote the historical momentum memory, the quantization operator for communication efficiency, and the residual correction term for drift mitigation, respectively.

\subsection{A-FedPD~\cite{sun2024fedpd}} 
While proposes primal-dual optimization frameworks that focus on theoretically aligning dual-drift to guarantee convergence under unbounded dynamics, they often introduce sophisticated constraints that complicate the FL optimization trajectory. 
\begin{equation}
    \nabla F_{a-pd} =  \nabla F_{r,k} + \lambda(\theta_{r,k} - \theta_r^g) + \tilde{\lambda}_{r,k}.\label{eq:afedpd}
\end{equation}
where, $\lambda$ denotes the penalty parameter for the proximal primal dual alignment, and $\tilde{\lambda}_{r,k}$ represents the aligned virtual dual variable utilized to mitigate unaligned dual-drift across heterogeneous clients.

\subsection{Summary of Limitations and Challenges}
To mitigate client drift induced by statistical heterogeneity, classical frameworks like FedProx~\cite{fedprox} introduce local proximal terms to restrict client updates. Similarly, adaptive optimization strategies such as FedOpt~\cite{reddi2020adaptive} employ server-side momentum to smooth global trajectories. Recent advanced frameworks, including AdaBest~\cite{varno2022adabest} and FedEve~\cite{shen2025fedeve}, apply dynamic bias estimation and predict-observe filtering to stabilize client and period drift.
However, a fundamental limitation of these approaches lies in their reliance on rigid, ad-hoc regularization heuristics applied directly to local updates. Under extreme unbounded drift and non-IID conditions, forcing client updates to strictly align with immediate global constraints can distort the true empirical likelihood of local data. This structural restriction introduces a severe trade-off between drift mitigation and global model capacity, ultimately resulting in suboptimal generalization and slower empirical convergence when encountering volatile user dynamics.

\section{FedRef: Bayesian Fine-tuning Using a Reference Model}

To alleviate unbounded drift, FedRef uses Bayesian fine-tuning with a reference model constructed from previous global models. It calibrates unstable global updates through a Bayesian transfer learning–based fine-tuning procedure. Moreover, to address the client-side computational bottleneck, our method assigns all Bayesian optimization overhead to the server. Figure~\ref{fig:federatedlearningsystem} illustrates our whole FedRef system.

\subsection{Client-Side Local Training}
As illustrated in Figure~\ref{fig:federatedlearningsystem}, FedRef framework decouples the probabilistic drift-control mechanism from the client-side optimization process. 
During the local training phase of communication round $r$, each selected client $k \in K$ independently minimizes its empirical risk utilizing its local dataset $D_k$. The generalized local optimization update can be formulated as follows:
\begin{equation}
    \theta_{r,k}^{s+1} = \theta_{r,k}^s - \eta \cdot \nabla Opt(\theta_{r,k}^s; \mathcal{B}_k^s),
\end{equation}
where $Opt$ represents the client-side optimization function, such as Equations (\ref{eq:adabest}), (\ref{eq:fedeve}), (\ref{eq:afedpd}), which can be flexibly configured using standard empirical loss or advanced local optimization frameworks without server-side dependency.

\subsection{Server-Side Reference Model Estimation}

Rather than computing a simple static average of past parameters, FedRef dynamically synthesizes the reference model $\theta^{ref}$ by incorporating a temporal proximity weighting, computed as: 
\begin{equation}
\theta_{r+1}^{ref} = \text{A}(\theta^g_{r-p}, \theta^g_{r-p+1}, \dots, \theta^g_r).
\label{eq:refmodel}
\end{equation}
This design mitigates the influence of erratic updates or early under-fitted global models, creating a smoother and more reliable optimization path. Here, the weighted aggregation function $A(\forall\theta)$ is computed as:
\begin{equation}
\text{A}(\forall\theta ) =
\sum_{i=1}^\rho  \frac{\rho-i+1}{\phi}\theta^g_{r+1-i}.
\label{eq:aggreation}
\end{equation}
where $\phi= \sum_{i=1}^\rho$ represents the normalization factor. By assigning exponentially higher structural weight to temporally adjacent rounds, $\theta^{ref}$ captures the long-term evolutionary trend of the global optimization trajectory. This formulation provides a robust probabilistic anchor that attenuates transient noise and severe data poisoning effects, thereby driving faster convergence and superior generalizability across diverse real-world tasks.

\subsection{Bayesian Transfer Learning Optimization}
We adopt a Bayesian transfer learning formulation \cite{bayesian} to mitigate unbounded drift by calibrating model updates through Maximum A Posteriori (MAP) estimation, as follows:
\begin{equation}
\theta^{*} = \arg\max_{\theta} \left[ \log p(D_B|\theta) + \log p(\theta|D_A) \right],
\end{equation}
where $\theta^{*}$ is an optimal model to address the specific task, $D_A$ and $D_B$ represent the data used for pretraining and the data used for downstream fine-tuning, respectively. Here, $\log p(\theta|D_A)$ can be approximated by Laplace estimation \cite{mackay1992practical} as:
\begin{equation}
\log p(\theta|D_A) \approx f(\theta_0) + \tfrac12 (\theta - \theta_0)^\top H (\theta - \theta_0),\label{eq:laplace}
\end{equation}
where $H$ is the negative Hessian of the log-posterior, approximated by the Fisher Information Matrix (FIM).

Finally, the minimized cost function for the optimization of transfer learning is defined in \cite{bayesian} as follows:
\begin{equation}
F(\theta) = \underbrace{F_B(\theta)}_{\text{task loss}} + \lambda(\theta - \theta_0)^\top \text{FIM}(\theta - \theta_0),
\label{eq:objectf}
\end{equation}
where $\lambda$ denotes the regularization strength. According to the Bayesian transfer learning study presented in \cite{bayesian}, the simplified FIM effectively minimizes the computational cost of fine-tuning as follow:
\begin{equation}
F_{\text{l2-sp}}(\theta) = F_B(\theta) + \lambda  \|\theta_k - \theta_{0,k}\|^2.
\label{eq:l2sp}
\end{equation}

Even with this simplification, the approach has been shown to improve robustness of model updates and to alleviate unbounded drift updates in transfer learning \cite{xuhong2018explicit, kirkpatrick2017overcoming}. Under alternative assumptions, Kronecker-factored estimation can also be employed \cite{martens2015optimizing, george2018fast}, as it accounts for parameter covariance within each layer.
To motivate the proposed formulation, we investigate a Bayesian transfer learning–based regularization term for FL scenarios to address unbounded drift in model updates, which arise from data diversity, data heterogeneity or malicious or misleading client behavior.

Bayesian parameter-efficient fine-tuning provides a rigorous probabilistic foundation to mitigate catastrophic forgetting and preserve generalization during downstream tasks~\cite{bayesian}. By formulating the optimization objective through Maximum A Posteriori (MAP) estimation, prior knowledge embedded in a pretrained model can be effectively preserved via Laplace approximation ~\cite{mackay1992practical, kirkpatrick2017overcoming}.
While conventional transfer learning operates on a static source-to-target paradigm ~\cite{bayesian}, federated learning demands a continuous calibration mechanism capable of handling aggregated, non-IID trajectories.

\subsection{Server-Side Bayesian Fine Tuning}
Our Bayesian fine tuning reduces the dual-drift arising from unbounded dynamics in model updates. Furthermore the server-side process can motivate clients to participate as high-performing contributors. Even though, clients set diverse hyperparameters causing dual-drift (e.g. learning rate, epoch per round, etc\dots), Bayesian fine tuning can calibrate model updates using a reference model generalized from prior global models. It not only consider last temporal unbounded drifts, but also address temporal unbounded drifts from $\rho$ rounds temporal unbounded drifts.

\subsubsection{MAP Formulation in Federated Learning}
Existing federated learning optimization frameworks often introduce strict, handcrafted regularization terms into local training objectives, which inevitably distorts the empirical client likelihood and restricts global model capacity under severe non-IID data distributions. To circumvent this generalization barrier, FedRef preserves the integrity of each client’s local likelihood optimization while establishing a comprehensive Maximum A Posteriori (MAP) framework at the global aggregation phase. Under this formulation, the global optimization target $\hat\theta$ is defined as follows:
\begin{equation}
\hat\theta = \arg\max_{\theta \in \theta^g} \left( \log p( \theta|D_{ref})+\sum_k^K{\log p(D_{k}|\theta)} \right),
\label{eq:flmap}
\end{equation}
where $D_{ref}$ denotes a conceptual reference dataset containing the temporally accumulated knowledge from preceding global trajectories, $\sum_k^K{\log p(D_{k}|\theta_k)}$, represents the joint likelihood across participating clients. By decoupling the prior alignment from the local training process, FedRef ensures that the global model aggregates raw empirical gradients from decentralized tasks while applying a robust probabilistic calibration that effectively counteracts unbounded drift under extreme system dynamics.

\subsubsection{Empirical Risk Minimization and Gradient Descent}
Empirical risk minimization (ERM) is used to find the best performing model in machine learning by minimizing the average error defined as empirical risk calculated on a finite train dataset. To connect between the our MAP formulation and ERM principle, we denoted as follows:
\begin{equation}
\hat \theta = \arg\min_{\theta \in \theta^g}\mathbb{E} \left ( F(y, \log p(D | \theta)) \right),
\label{eq:erm}
\end{equation}

Based on the MAP formulation, we design an objective function for target-model fine-tuning. When unbounded drift occurs, this objective guides the global model toward a bounded and temporally stable parameter region.
To incorporate the client loss function, we can formulate the objective cost function representing model predictive error computed via log posteriori on the clients test samples, as follows:
\begin{IEEEeqnarray}{rCl}
F_k(y_k, \log p(D_k | \theta_k) = - \log p(D_k\mid\theta_k),\nonumber\\ \hat \theta = \arg\min_{\theta \in \theta^g}\left(\underbrace{\sum_{k=1}^K  W_kF_k   + F_{ref}(\theta | D_{ref})}_{f: Bayesian Fine Tuning}\right).
\label{eq:fedref}
\end{IEEEeqnarray}
where $F_k$ represents the local loss value of client $k$,
$W_k$ denotes a client $k$ aggregation weights (e.g., $\frac{n_k}{n}$).
To leverage a reference model to help alleviate unbounded drift in model updates, FedRef employed a reference model estimation approach using prior global models. In our MAP estimation, using a Laplace approximation, the prior global models contribute to the computation of a novel $L_2$ regularization term that encourages the updated global model to remain close to the reference model. This helps alleviate unbounded drift by penalizing updates that deviate excessively from the reference model in heterogeneous FL environments.
Finally, FedRef can fine tune and calibrate global model updates with the log-posteriori estimation from Eq.~\ref{eq:flmap}-\ref{eq:fedref} and the prior regularization from Eq.~\ref{eq:l2sp}. This cost function can balances the drift of the current global model against the reference-model constraint. Our main cost function is defined as follows:
\begin{equation}
  F_{ref}(\theta| D_{ref}) = \lambda^g\|\Delta\theta^g_r
-  \Delta\theta^{ref}_r\|^2,
\label{eq:objectref}
\end{equation}
where $\theta^g$ denotes the global model, and $\theta^{ref}$ denotes the reference model. Here, the term $\|\Delta\theta^g_r
-  \Delta\theta^{ref}_r\|^2$ corresponds to an $L_2$ regularization on the difference between successive model parameters, $\theta_g$ and $\theta_{ref}$.
The server computes the gradient of $F_{ref}$ and leverages a single Bayesian fine-tuning to aggregated global model updates using the aggregated client losses $F_k$ and the reference constraints imposed by $\theta^{ref}$. This yields $\hat\theta$ as the optimized target model via simple gradient descent, computed as:
\begin{equation}
\hat\theta \leftarrow \theta^g_r-\eta\times\nabla{f}(\theta^g_r,\theta^{ref}_r, F^g_r, \lambda^g),
\label{eq:gradientdecent}
\end{equation}
where $\eta$ represents the learning rate. This procedure is performed on the server side to fine-tune the model and calibrate global updates.
\begin{algorithm}
\caption{FedRef: Bayesian fine-tuning with a reference model}
\label{alg:fedref}
\begin{algorithmic}[1]
\FOR {round $r$}
\STATE\textit{Client Operation:}
\FOR {epoch $e$}  
    \FOR {batch $s$}
         \STATE$\theta_{r,k}^{s+1} = \theta_{r,k}^{s} - \eta \nabla F_k(\theta_{r,k}^{s}; \mathcal{B}_k^{s})$ (Eq.~\ref{localtraining})
    \ENDFOR
\ENDFOR
\STATE $ F_k\leftarrow\sum_{s=1} F_k(\theta^{s}_{r,k}; \mathcal{B}^{s}_k)$
\STATE Send $\theta$ ,  $ F_k$ to Server\\
\STATE\textit{Server Operation:}

\IF {$r\leq\rho$ :}
  \STATE    $\textbf{FedAveraging}()$ (Eq.~\ref{averaging})
  \STATE \text{Continue;}
\ENDIF
\STATE  $K$-client selection
\STATE  $F^g_{r+1}\leftarrow\sum_{i=1}^K W_i F_i, \quad i\in K$
\STATE  \IF {$\Delta F^g_{r+1} < \delta$} 
\STATE    $\textbf{FedAveraging}()$ (Eq.~\ref{averaging})
\STATE    Continue;
        \ENDIF
\STATE  $\theta^g_{r+1}$ $\leftarrow$$\textbf{Aggregation} (\theta_{r,1}, \theta_{r,2}, \ldots,\theta_{r,K})$ 
\STATE  $\theta^{ref}_{r+1}$ $\leftarrow$ $\textbf{Estimation} (\theta^g_{r-p},\ldots,\theta^g_{r-1},\theta^g_{r})$ (Eq.~\ref{eq:refmodel},\ref{eq:aggreation})
\STATE  $\hat\theta$ $\leftarrow$  $\textbf{Bayesian Fine-Tuning} (\theta^g_{r+1}, \theta^{ref}_{r+1}, F^g_{r+1}, \lambda^g)$ 
\STATE  $\theta^g_{r+1} \leftarrow \hat{\theta}$
\STATE  \textbf{Broadcast} $\theta^g_{r+1}$ to Clients
\ENDFOR \\
\STATE \textbf{Aggregation ($\forall \theta_k)$}:  
\STATE \quad \textbf{return} $\sum_{k=1}^K \frac{n_k}{n} \theta_{k}$\\
\STATE\textbf{Estimation $(\theta^g_{r-p},\ldots,\theta^g_{r-1},\theta^g_{r})$}: 
\STATE \quad \textbf{return} $\sum_{i=1}^\rho  \frac{\rho-i+1}{\phi}\theta^g_{r+1-i}$ \\
\STATE\textbf{Bayesian Fine-Tuning ($\theta^g, \theta^{ref},  F^g, \lambda^g$)}:~(Eq.~\ref{eq:fedref}-\ref{eq:gradientdecent})
  \STATE \quad $\hat\theta \leftarrow \theta^g-\eta\times\nabla{F}_{ref}(\theta^g,\theta^{ref},F^g, \lambda^g)$
 \STATE \quad \textbf{return} $\hat\theta$\\
    
\STATE \textbf{Output} \,$\theta^g_R$
\end{algorithmic}\end{algorithm}
To illustrate FedRef whole algorithm, in Algorithm~\ref{alg:fedref} (lines 2-8), each client $k$ trains the global model $\theta_{r}^{g}$ using its local data to obtain an updated local model $\theta_{k}^{s}$, and then sends the final local model $\theta$ along with the average cost value $F(\theta^{s}; \mathcal{B}^{s})$ to the server. On the server side (lines 9–27), FedRef performs Bayesian fine-tuning to mitigate unbounded drift, using the aggregated global model and the reference model and the client loss values received from participating clients.

By utilizing the raw empirical client loss values $F_k$ as proxies for the negative log-likelihood~\ref{eq:fedref}, FedRef circumvents the need for complex, resource-intensive weight parameter adjustments during local updates. This strategic formulation guarantees a smooth objective function $F_{ref}$ that balances immediate empirical likelihood against long-term temporal priors. Furthermore, as illustrated in Algorithm~\ref{alg:fedref}, our framework avoids continuous underfitting by incorporating a dynamic threshold conditional trigger $(\Delta F^g_{r+1}> \delta)$. Bayesian fine-tuning is only deployed when severe unbounded drift is detected, thereby sustaining high optimization velocity under standard iterations while providing robust performance margins under highly volatile environments.

\section{Experiments}

\subsection{Environmental Setup}
Our experiments evaluate multi‑class image classification and semantic segmentation of medical images. For classification, we use FEMNIST and CINIC‑10 as benchmark datasets. For medical image segmentation, we adopt the FeTS2022 dataset, which is partitioned under a distributed non‑IID configuration.
To assess predictive performance, we adopt classification metrics such as accuracy and F1-score, as well as segmentation metrics commonly used in medical image analysis~\cite{karimi2019reducing, chen2021transunet, hatamizadeh2022unetr, zhu2022medical}, including mean Intersection over Union (mIoU), Dice coefficient (DC), and Hausdorff distance at the 95th percentile (HD95).
For demonstration on better model convergence, client‑side computational efficiency, and alleviating unbounded drift penalty, we compare FedRef against standard federated optimization baselines to alleviate unbounded drift model updates, including FedAvg~\cite{fedavg}, FedProx~\cite{fedprox}, FedOpt~\cite{reddi2020adaptive}, AdaBest~\cite{varno2022adabest}, FedEve~\cite{shen2025fedeve}. 

We implemented our federated learning experiments using the Flower framework~\cite{beutel2020flower}. As shown in Table~\ref{tab1:values}, we set the hyperparameters of each baseline according to the optimal settings reported in the corresponding paper, using the values that yielded the best predictive performance for each dataset.

\begin{table}[t]
\centering
\caption{Hyperparameters}
\label{tab1:values}

\begin{tabular}{l l}
\toprule
\textbf{Hyperparameter} & \textbf{Value} \\
\midrule
Number of clients ($K$) & 10 \\
Number of epochs per round & 3 \\

FedProx (\ref{eq:fedprox}) & $\mu = 0.5$ \\
FedOpt (\ref{eq:fedopt}) & $\beta_1=0.9$, $\beta_2=0.99$, $\tau=10^{-4}$ \\
AdaBest (\ref{eq:adabest}) & $\alpha:0.1$ , $\beta :1.0$ \\
FedEve (\ref{eq:fedeve})& $\beta:0.9 $, $\mu:0.01 $, $\rho:0.9 $ \\
A-FedPD (\ref{eq:afedpd})& $\mu: 0.1 $ \\
FedRef (\ref{eq:fedref})& $\lambda^g : 10^{-3}$, $\rho \in [1,2,4,8]$, $\delta : [-f, f]$ \\
\bottomrule
\end{tabular}
\end{table}


\begin{figure*}[!thb]
\centering
\includegraphics[width=0.32\textwidth]{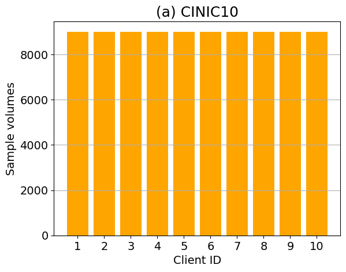}
\includegraphics[width=0.32\textwidth]{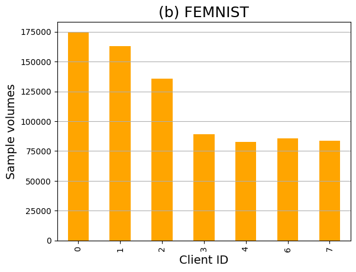}
\includegraphics[width=0.32\textwidth]{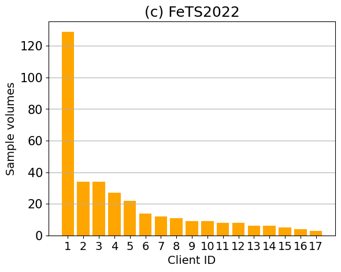}
\caption{Number of samples per client for each dataset: CINIC-10 (high IID), FEMNIST (low non-IID), and FeTS2022 (high non-IID).}
\label{fig:partitioning}
\end{figure*}

\subsection{Datasets}
As shown in Figure~\ref{fig:partitioning}, we use three dataset varying degrees of data heterogeneity: FEMNIST, CINIC-10, and FeTs2022.
Figure~\ref{fig:partitioning} illustrates the number of samples allocated to each client index. The experimental settings for these three datasets exhibit different degrees of data skew, reflecting different levels of client heterogeneity. In contrast, the CINIC-10 setting has a relatively homogeneous data distribution. FEMNIST exhibits a more heterogeneous data distribution than CINIC-10, whereas FeTS2022 shows the most severe heterogeneity among the three datasets.
\subsubsection{CINIC10}
It is a large-scale image classification dataset designed as a drop-in replacement for CIFAR-10 in FL research. It was created to address the need for a more challenging benchmark dataset that fills the gap between the relatively small CIFAR-10 dataset and the much larger ImageNet dataset~\cite{DBLP:journals/corr/abs-2007-14390, darlow2018cinic10imagenetcifar10}. For the multi-class classification task, we adopt the asymmetric loss~\cite{ridnik2021asymmetric}, which imposes asymmetric penalties on over and under estimated predictions to better model directional deviations from the ground truth. Under the high IID CINIC10 dataset, the asymmetric reducing class imbalance provides improved performance compared to cross-entropy.
To design heterogeneous experimental settings, we distribute data samples to each client under the same volumes, but can cause and exhibit class imbalance problems.

\subsubsection{FEMNIST}
It is an image classification benchmark consisting of handwritten digits (0-9), lowercase letters (a-z), and uppercase letters (A-Z). This results in 62 unique labels.
The dataset is available through the Hugging Face Hub, making it easy to integrate with the Flower federated learning framework~\cite{DBLP:journals/corr/abs-1812-01097, DBLP:journals/corr/abs-2007-14390}. The local training settings are the same as those used for CINIC10. But data distribution represents more heterogeneous and skew while using the hsf-id information included in FEMNIST dataset.

\subsubsection{FeTS2022} 
It was introduced in the Federated Tumor Segmentation (FeTS) Challenge 2022, which focused on brain tumor segmentation using FL approaches~\cite{fets2022}. The dataset contains multimodal brain MRI scans of glioma patients.

To evaluate the brain tumor segmentation model, on the FeTs2022 dataset under a specific non-IID data distribution setting, we adopt 3D U-Net~\cite{cciccek20163d} as a baseline model and, in the local training settings, we employ the focal Dice loss~\cite{lin2017focal, yeung2022unified}, which applies difficulty‑weighted penalties to poor overlap regions, emphasizes hard samples, and mitigates the severe class imbalance prevalent in medical imaging. To mitigate the brain tumor class imbalance under the dataset, we employ the focal Dice loss under the heterogeneous medical federated learning scenarios.

\subsection{Heterogeneous and Dual-Drift Settings}
Existing works\cite{varno2022adabest,shen2025fedeve} categorized dual drift to period drift representing deviation (between participating clients' model and global model) and client drift representing deviation (among participating clients' models). Period drift often arise from disturbances of global model aggregation (communication failures, uploading zero updates including partial participation, free-rider attacks). Client drift often arise from (diversity of local training settings including multiple local steps or learning rates, data heterogeneity, data noisy).

To empirically simulate the mathematical challenges of unbounded dynamics and relaxed smoothness in real-world federated learning, we establish a volatile optimization environment driven by two realistic system-level hazards: step-size heterogeneity and asynchronous communication delays. 

First, to introduce severe optimization trajectory mismatch, we implement client-side learning rate heterogeneity. Instead of utilizing a uniform step size, each participating client is assigned a distinct local learning rate randomly sampled from a highly volatile search space $\eta_k \in \{10^{-2}, 10^{-3}, 10^{-4}, 9\times10^{-6}, 4\times10^{-6}, 10^{-6}\}$. This variation forces volatile clients causing extreme model overfitting or underfitting during local iterations, successfully maximizing the structural client drift. 

Second, we simulate temporal network instability by injecting stale updates (stragglers) during the server-side aggregation phase. Specifically, we define the Unbounded Drift Probability (UDP) as the operational threshold that determines the injection rate of temporal anomalies. With a probability defined by UDP (ranging from 0.0 to 1.0), the updates of randomly selected clients are artificially substituted with stale global model parameters from previous communication rounds (i.e., $\theta_{r-\tau}$ where $\tau \in \{1, 2, 3\}$). A UDP of 0.0 signifies a perfectly synchronous environment with a static learning rate, whereas a UDP of 1.0 represents a continuously chaotic optimization landscape where every round suffers from severe dual-drift and non-continuous gradient trajectories. 

By compounding these two structural perturbations, the consecutive global trajectories become structurally unaligned, thereby preventing the optimization trajectory from being bounded and providing a rigorous testbed for our FedRef framework.
\subsection{Model Divergence Quantification Method}
To quantify unbounded drift in model updates and model divergence, Kullback-Leibler Divergence (KLD) often classically be used to assess model updates representing model divergence or unbounded drift. We adopt KLD to quantify decrease of unbounded drift in model updates. The KLD formulation computed as:
\begin{equation}
    {\text{KLD}} (f_{out}^1 ||f_{label}^2) = \sum_{x \in \mathbf{D(x)}}f^1_{out}(x) \log \frac{f^1_{out}(x)}{f^2_{label}(x)} 
    \label{eq:KLD}
\end{equation}
where $f^1_{out}$ : target model yield probability, $f^2_{label}$ : standard label values using label logits in dataset $D_k$. Low KLD represent low unbounded drift in the model update. In the other hand, high KLD represent high unbounded drift in the model updates and model divergence.
\begin{equation}
    f^2_{label} =\begin{cases}
       1.0 & \text{if y = 1} \\
       10^{-n} & \text{if y = 0} 
    \end{cases}
    \label{eq:f2}
\end{equation}
Here, the Equation (\ref{eq:f2}) prevent zero division errors and KLD divergence in Equation (\ref{eq:KLD}).
\subsection{Experimental Results}
\begin{table}
\centering
\caption{FedRef Hyperparameter Sensitivity in CINIC-10}
\label{tab:cinic10:hypersensitivity}
\begin{tabular}{c|c|c|c|c|c}
\toprule
UDP & $\lambda$ & $\rho$ & KLD & Accuracy ($\%$) & F1score ($\%$)\\
\midrule
\multirow{9}{*}{0.2}&\multirow{3}{*}{0.01}& 1&\multirow{9}{*}{2.050}& 25.09&25.67 \\
&& 4&& 25.11&25.68\\
&& 8&& 25.06&25.64\\
&\multirow{3}{*}{0.1}& 1& & 25.13 & 25.76 \\
&& 4& &25.08 &25.66\\
&& 8& & 25.08&25.69\\
&\multirow{3}{*}{0.9}& 1&& 25.09& 25.76\\
&& 4& & \textbf{25.13}& \textbf{25.79}\\
&& 8&& 25.05&25.65\\ \midrule
\multirow{9}{*}{0.4}&\multirow{3}{*}{0.01}& 1& \multirow{9}{*}{2.104} & 22.73& 23.23\\
&& 4& &22.76& 23.27\\ 
&& 8& &22.81& 23.37\\ 
&\multirow{3}{*}{0.1}& 1& & 22.76&23.30\\
&& 4& &22.80& 23.32\\
&& 8& &22.78& 23.31\\
&\multirow{3}{*}{0.9}& 1& & 22.73&23.25\\
&& 4& & 22.81&23.32\\
&& 8& & \textbf{22.82}&\textbf{23.93}\\ \midrule
\multirow{9}{*}{0.6}&\multirow{3}{*}{0.01}& 1& \multirow{9}{*}{2.36}& \textbf{12.75} & 18.09\\
&& 4& & 12.71&18.08\\ 
&& 8& & 12.72&18.09\\ 
&\multirow{3}{*}{0.1}& 1& & 12.71&\textbf{18.10}\\
&& 4& & 12.71& 18.08\\
&& 8& & 12.73 & 18.09\\
&\multirow{3}{*}{0.9}& 1& & 12.73&18.09\\
&& 4& & 12.71&18.08\\
&& 8& & 12.72&18.09\\ \bottomrule
\end{tabular}
\end{table}
As presented in Table~\ref{tab:cinic10:hypersensitivity}-\ref{tab:fets:hypersensitivity}, our extensive evaluations demonstrate that the proposed FedRef framework consistently outperforms state-of-the-art baselines under severe non-IID and adversarial settings. The hyperparameter sensitivity analysis reveals that both the unbounded drift penalty parameter $\lambda^g$ and the history window length $\rho$ exert a direct and profound influence on model performance. Specifically, higher values of $\lambda$ and $\rho$ strengthen the Bayesian fine-tuning process, yielding superior predictive accuracy compared to lower configurations by establishing a stable temporal probabilistic anchor.

Across both the homogeneous CINIC-10 and highly heterogeneous FEMNIST classification benchmarks, FedRef consistently achieves lower KLD values alongside higher accuracy and F1-scores relative to traditional methods such as FedAvg, FedProx, FedOpt, AdaBest, FedEve, and A-FedPD. Even under elevated Unbounded Drift Probability (UDP) conditions, such as $0.4$ or $0.6$, FedRef demonstrates remarkable resilience against systemic volatility, delayed updates, and communication artifacts. Furthermore, on the challenging FeTS2022 brain tumor segmentation benchmark, FedRef successfully maintains high Dice coefficients while significantly reducing the Hausdorff distance. By decoupling probabilistic drift control from local client updates and executing server-side Bayesian fine-tuning via temporally aggregated reference models, FedRef effectively suppresses unbounded drift without imposing excessive computational overhead on decentralized clients.

\begin{table}
\centering
\caption{FedRef Hyperparameter Sensitivity in FEMNIST}
\label{tab:femnist:hypersensitivity}
\begin{tabular}{c|c|c|c|c|c}
\toprule
UDP & $\lambda$ & $\rho$ & KLD & Accuracy ($\%$) & F1score ($\%$)\\
\midrule
\multirow{9}{*}{0.2}&\multirow{3}{*}{0.01}& 1&3.987& 65.30&70.62 \\
&& 4&3.992& 65.25&70.61\\
&& 8&3.991& 65.31&70.62\\
&\multirow{3}{*}{0.1}& 1& 3.975 & \textbf{65.38} & \textbf{70.69} \\
&& 4& 4.002& 65.17 &70.50\\
&& 8& 3.974& 65.31&70.62\\
&\multirow{3}{*}{0.9}& 1&3.991& 65.32& 70.67\\
&& 4& 3.992& 65.24& 70.59\\
&& 8&3.992& 65.25&70.58\\ \midrule
\multirow{9}{*}{0.4}&\multirow{3}{*}{0.01}& 1& 4.26 & \textbf{62.33}& \textbf{67.97}\\
&& 4& 4.26&61.90& 67.61\\ 
&& 8& 4.48&61.35& 67.19\\ 
&\multirow{3}{*}{0.1}& 1& 4.28& 61.86&67.51\\
&& 4& 4.29&62.06& 67.74\\
&& 8& 4.27&62.20& 67.90\\
&\multirow{3}{*}{0.9}& 1& 4.25& 62.03&67.73\\
&& 4& 4.46& 61.73&67.56\\
&& 8& 4.27& 62.07&67.75\\ \midrule
\multirow{9}{*}{0.6}&\multirow{3}{*}{0.01}& 1& 7.364 & \textbf{28.72} & \textbf{36.83} \\
&& 4& 7.302&28.13&36.12\\ 
&& 8& 7.412&28.01&35.98\\ 
&\multirow{3}{*}{0.1}& 1& 7.366& 28.38&36.39 \\
&& 4& 7.414&27.95& 35.93\\
&& 8& 7.380&27.88& 35.86\\
&\multirow{3}{*}{0.9}& 1& 7.396& 27.57&35.48\\
&& 4&7.396&27.79 &35.36\\
&& 8& 7.396& 27.99&35.78\\ \bottomrule
\end{tabular}
\end{table}

\begin{table}
\centering
\caption{FedRef Hyperparameter Sensitivity in FeTS}
\label{tab:fets:hypersensitivity}
\begin{tabular}{c|c|c|c|c|c}
\toprule
UDP & $\lambda$ & $\rho$ & KLD & Dice Coefficient ($\%$) & Hausdorff Distance\\
\midrule
\multirow{9}{*}{0.2}&\multirow{3}{*}{0.01}& 1& 6556286 & \multirow{9}{*}{32.39}&10.51 \\
&& 4&6556006& &9.19\\
&& 8&6562821& &10.54\\
&\multirow{3}{*}{0.1}& 1& 6558238&  & 9.51 \\
&& 4& 6553981& &11.24\\
&& 8& 6550471& &\textbf{8.43}\\
&\multirow{3}{*}{0.9}& 1&6562045& & 8.60\\
&& 4&6561219 & & 11.19\\
&& 8&6554575& &8.62\\ \midrule
\multirow{9}{*}{0.4}&\multirow{3}{*}{0.01}& 1& 6553155 &\multirow{9}{*}{32.39}& 9.32\\
&& 4& 6561240&& 9.11\\ 
&& 8& 6565174&& 10.54\\ 
&\multirow{3}{*}{0.1}& 1& 6557522& &9.16\\
&& 4&6554714 &&9.58\\
&& 8& 6553792&& 9.88\\
&\multirow{3}{*}{0.9}& 1& 6555959& &\textbf{8.67}\\
&& 4& 6553295& &9.28\\
&& 8& 6553731& &9.65\\ \midrule
\multirow{9}{*}{0.6}&\multirow{3}{*}{0.01}& 1& \multirow{9}{*}{6553077}& \multirow{9}{*}{31.29} & \multirow{9}{*}{9.17}\\
&& 4& & &\\ 
&& 8& & &\\ 
&\multirow{3}{*}{0.1}& 1& & & \\
&& 4& & & \\
&& 8& &  & \\
&\multirow{3}{*}{0.9}& 1& & &\\
&& 4& & &\\
&& 8& & &\\ \bottomrule
\end{tabular}
\end{table}

\begin{table}
    \centering
    \begin{tabular}{c|c|c|c|c|c}
    \toprule
        Dataset & Method & UDP & KLD & Accuracy (\%) & F1score (\%)\\
    \midrule
        \multirow{7}{*}{CINIC-10} & FedAvg & \multirow{7}{*}{0.2} & 2.050 &25.13&25.72\\ 
         & FedProx &  & 2.050 &25.09&25.71\\ 
         & FedOpt &  &  2.050&25.09&25.73\\ 
         & Adabest &  &  2.325&13.82&14.70\\ 
         & FedEve &  &  2.064&24.50&25.12\\ 
         & A-FedPD &  &  2.051&25.09&25.72\\ 
         & \textbf{FedRef} &   & 2.050 & \textbf{25.13}& \textbf{25.79}\\ 
    \midrule
        \multirow{7}{*}{CINIC-10} & FedAvg & \multirow{7}{*}{0.4} & 2.104 &22.76&23.29\\ 
         & FedProx &  &  2.104&22.78&23.31\\ 
         & FedOpt &  &  2.104&22.77&23.28\\ 
         & Adabest &  &  2.427&12.00&16.52\\ 
         & FedEve &  &  2.119&22.19&22.73\\ 
         & A-FedPD &  &  2.105&22.77&23.33\\ 
         & \textbf{FedRef} &   & 2.104 &\textbf{22.82}& \textbf{23.93} \\ 
    \midrule
        \multirow{7}{*}{CINIC-10} & FedAvg & \multirow{7}{*}{0.6} & 2.361 &12.70&18.09\\ 
         & FedProx &  & 2.361 &12.72&18.09\\ 
         & FedOpt &  & 2.361 &12.69&18.09\\ 
         & Adabest &  & 2.388 &10.53&18.00\\ 
         & FedEve &  &  2.361&12.68&\textbf{18.34}\\ 
         & A-FedPD &  &  2.360&12.72&18.09\\ 
         & \textbf{FedRef} &  & 2.36 &\textbf{12.75}& 18.09\\ 
    \bottomrule
    \end{tabular}
    \caption{Predictive Performance Table in Cinic-10}
    \label{tab:all:classification:performance}
\end{table}

\begin{table}
    \centering
    \begin{tabular}{c|c|c|c|c|c}
    \toprule
        Dataset & Method & UDP & KLD & Accuracy (\%) & F1score (\%)\\
    \midrule
        \multirow{7}{*}{FEMNIST} & FedAvg & \multirow{7}{*}{0.2} & 3.995 & 65.32&70.65\\ 
         & FedProx &  & 3.989 &65.32&70.64\\ 
         & FedOpt &  &  3.994&65.27&70.60\\ 
         & Adabest &  &  5.101&35.68&43.04\\ 
         & FedEve &  &  3.197&61.75&66.42\\ 
         & A-FedPD &  &  4.095&64.81&70.24\\ 
         & \textbf{FedRef} &  & 3.975 &\textbf{65.38}&\textbf{70.69}\\ 
    \midrule
        \multirow{7}{*}{FEMNIST} & FedAvg & \multirow{7}{*}{0.4} & 4.31 &61.74&67.44\\ 
         & FedProx &  & 4.290 &61.93&67.65\\ 
         & FedOpt &  &  4.299&61.76&67.46\\ 
         & Adabest &  & 7.240 &8.36&11.71\\ 
         & FedEve &  &  4.244&60.82&66.58\\ 
         & A-FedPD &  &  4.310&61.50&67.25\\ 
         & \textbf{FedRef} &  & 4.26 & \textbf{62.33}& \textbf{67.97}\\ 
    \midrule
        \multirow{7}{*}{FEMNIST} & FedAvg & \multirow{7}{*}{0.6} &  7.317&28.43&36.49\\ 
         & FedProx &  & 7.429 &27.17&35.02\\ 
         & FedOpt &  & 7.333 &28.47&36.54\\ 
         & Adabest &  & 13.84 &2.71&4.92\\ 
         & FedEve &  &  6.377&24.82&32.12\\ 
         & A-FedPD &  &  7.426&27.76&35.68\\ 
         & \textbf{FedRef} &  &7.364  &\textbf{28.72}&\textbf{36.83}\\ 
    \bottomrule
    \end{tabular}
    \caption{Predictive Performance Table in FEMNIST}
    \label{tab:all:classification:performance}
\end{table}

\begin{table}[]
    \centering
    \begin{tabular}{c|c|c|c|c|c}
    \toprule
    Dataset & Method & UDP & KLD & DC (\%)& Hausdorff Distance \\
    \midrule
        \multirow{7}{*}{FeTS} & FedAvg & \multirow{7}{*}{0.2} & 6561239 &32.39&10.02\\ 
         & FedProx &  & 6554945 &32.38&9.30\\ 
         & FedOpt &  & 6564171 &32.39&9.15\\ 
         & Adabest &  &  6564214&32.09&9.08\\ 
         & FedEve &  & 6550415 &32.34&11.32\\ 
         & A-FedPD &  & 6562726 &32.14&9.84\\ 
         & \textbf{FedRef} &  & 6550471 &32.39&\textbf{8.43}\\
    \midrule
        \multirow{7}{*}{FeTS} & FedAvg & \multirow{7}{*}{0.4} &6557845 &32.39&11.21\\ 
         & FedProx &  & 6554173 &32.38&9.67\\ 
         & FedOpt &  &  6559496&32.39&9.88\\ 
         & Adabest &  &  6553549&31.69&9.52\\ 
         & FedEve &  &  6550415&32.34&11.32\\ 
         & A-FedPD &  &  6560317&32.05&9.52\\ 
         & \textbf{FedRef} &  &  6555959&32.39&\textbf{8.67}\\
    \midrule
        \multirow{7}{*}{FeTS} & FedAvg & \multirow{7}{*}{0.6} &6553076  &\multirow{7}{*}{31.29}&\multirow{7}{*}{9.17}\\ 
         & FedProx &  &6553076  &&\\ 
         & FedOpt &  &  6553076&&\\ 
         & Adabest &  & 6553079 &&\\ 
         & FedEve &  &  6553078&&\\ 
         & A-FedPD &  &  6553076&&\\ 
         & FedRef &  & 6553077&&\\
    \bottomrule
    \end{tabular}
    \caption{Predictive Performance Table in Image Segmentation}
    \label{tab:all:segmentation:performance}
\end{table}

\begin{table*}
\caption{Comparison of Optimization Mechanisms and Generalization Attributes}
\label{table:comparison}
\centering
\begin{tabular}{l|c|c|c}
\hline
\textbf{Method} & \textbf{Optimization Paradigm} & \textbf{Drift Control Strategy} & \textbf{Generalization Framework} \\ \hline
FedAvg~\cite{fedavg} & Standard Empirical Risk Minimization & Raw Model Aggregation & No Prior Alignment \\ \hline
FedProx~\cite{fedprox} & Local Proximal Regularization & Static Norm Constraints & Immediate Global Anchor \\ \hline
FedOpt~\cite{reddi2020adaptive} & Server-Side Adaptive Optimization & Momentum-based Smoothing & Gradient-driven Tracking \\ \hline
AdaBest~\cite{varno2022adabest} & Client-Side Adaptive Bias Estimation & Dynamic Update Norm Constraints & Local Bias Compensation \\ \hline
FedEve~\cite{shen2025fedeve} & Predict-Observe Filtering & Variance Reduction Heuristics & Dual Drift Filtering \\ \hline
A-FedPD~\cite{sun2024fedpd} & Primal-Dual Optimization & Aligning Dual-Drift & Dual Drift Balancing \\ \hline
\textbf{FedRef (Ours)} & \textbf{Maximum A Posteriori (MAP) Estimation} & \textbf{Temporal Reference Calibration} & \textbf{Temporally Aggregated Prior} \\ \hline
\end{tabular}
\end{table*}

\section{Conclusion}
In this paper, we introduced FedRef, a rigorous and robust federated optimization framework designed to conquer the generalization barriers induced by extreme statistical heterogeneity and unbounded dynamics in federated learning. Moving away from traditional heuristics that enforce rigid, distortive constraints on client-side updates, FedRef establishes a formal Maximum A Posteriori (MAP) estimation paradigm at the global level. By dynamically synthesizing a temporally aggregated reference model from past global trajectories, our approach provides a resilient probabilistic anchor that systematically calibrates misleading local updates without compromising empirical model capacity.
Through extensive empirical evaluations on diverse benchmarks ranging from complex computer vision tasks (FEMNIST, CINIC-10) to real-world medical image segmentation (FeTS2022). we have demonstrated that FedRef consistently outperforms state-of-the-art baselines. It delivers superior predictive accuracy, and higher F1-scores under severe non-IID and adversarial settings. Crucially, our conditional Bayesian calibration mechanism guarantees high optimization fidelity, intervening precisely when severe unbounded drift threatens global stability.
For future work, the mathematical flexibility of the FedRef framework opens up compelling avenues for broader federated learning paradigms. We plan to explore the integration of our temporal reference prior with client-side regularization frameworks, investigating hybrid optimization structures to further elevate personalization and robustness in massively distributed, decentralized networks. Ultimately, we believe FedRef establishes a foundational step toward achieving scalable, high-fidelity generalization in highly volatile real-world FL applications.


\bibliographystyle{IEEEtran}
\bibliography{reference}
\begin{IEEEbiography}[{\includegraphics[width=1in,height=1.25in,clip,keepaspectratio]{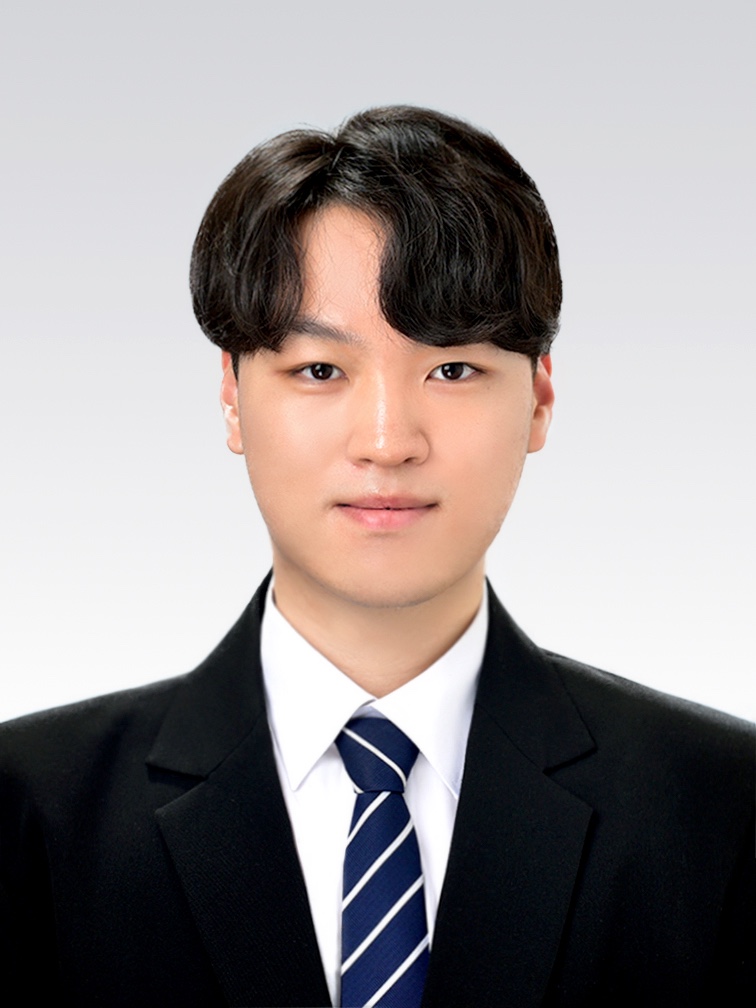}}]{Taehwan Yoon} received the B.S. degree from Korea Bible University, Seoul, South Korea, in 2023. He is currently pursuing the Ph.D. degree (M.Sc./Ph.D. joint program) in computer science and engineering with Soongsil University, Seoul, South Korea. Since 2023, he has been a Researcher with the Distributed Intelligence Laboratory (DILab) at Soongsil University. His research interests include artificial intelligence (AI), privacy-preserving machine learning (PPML), healthcare AI, and federated learning (FL).
\end{IEEEbiography}

\begin{IEEEbiography}[{\includegraphics[width=1in,height=1.25in,clip,keepaspectratio]{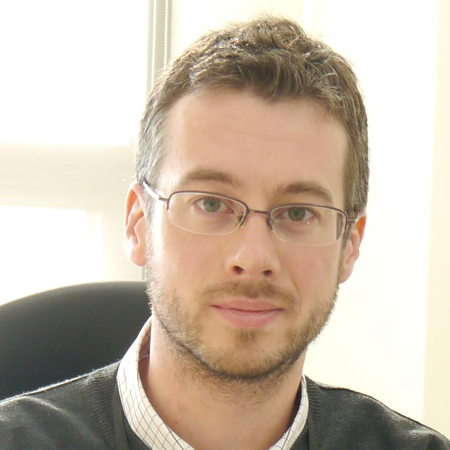}}]{Wesley De Neve} received the M.Sc. degree in Computer Science and the Ph.D. degree in Computer Science Engineering from Ghent University, Ghent, Belgium, in 2002 and 2007, respectively. He is currently working as an Associate Professor for both the IDLab at Ghent University in Belgium and the Center for Biotech Data Science at the Ghent University Global Campus (GUGC) in Korea. His teaching responsibilities are in the areas of Informatics (scientific problem solving using Python and UNIX) and Bioinformatics (algorithmic design and analysis at the intersection of computer science and biology). His research focuses on the development of novel compression and machine learning approaches towards the representation and analysis of multimedia content (that is, text and video) and biotech data (that is, omics data and medical images).
\end{IEEEbiography}

\begin{IEEEbiography}[{\includegraphics[width=1in,height=1.25in,clip,keepaspectratio]{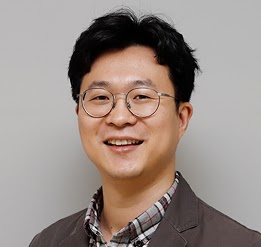}}]{David (Bong-Jun) Choi} 
Bong Jun Choi is an associate professor at the School of Computer Science $\&$ Engineering and jointly at the School of Electronic Engineering, Soongsil University, Seoul, Korea. Previously, he was an assistant professor at the Department of Computer Science, State University of New York Korea, Korea, and concurrently a research assistant professor at the Department of Computer Science, Stony Brook University, USA. He received his B.Sc. and M.Sc. degrees from Yonsei University, Korea, both in Electrical and Electronics Engineering, and his Ph.D. degree from the University of Waterloo, Canada, in Electrical and Computer Engineering. His current research focuses on distributed artificial intelligence, distributed intelligent energy networks, federated learning, and security. He is a senior member of IEEE and a member of ACM.
\end{IEEEbiography}

\end{document}